\pdfoutput=1

\documentclass[11pt]{article}

\usepackage{acl}
\usepackage{times}
\usepackage{latexsym}

\usepackage[T1]{fontenc}

\usepackage[utf8]{inputenc}

\usepackage{microtype}

\usepackage{amsmath,amsfonts,bm}

\def\eqref#1{equation~\ref{#1}}

\def\1{\bm{1}}

\DeclareMathAlphabet{\mathsfit}{\encodingdefault}{\sfdefault}{m}{sl}
\SetMathAlphabet{\mathsfit}{bold}{\encodingdefault}{\sfdefault}{bx}{n}

\usepackage{hyperref}
\usepackage{url}
\usepackage{inconsolata}
\usepackage{graphicx}
\usepackage{wrapfig}
\graphicspath{ {./Figs/} }
\usepackage[rightcaption]{sidecap}
\usepackage[T1]{fontenc}
\usepackage[utf8]{inputenc}
\usepackage{microtype}
\usepackage{pifont}
\usepackage{latexsym}

\usepackage{balance}
\usepackage{helvet}  
\usepackage{courier}  
\usepackage{url}  
\usepackage{graphicx}  
\usepackage{amssymb}
\usepackage{amsmath}
\usepackage{algorithm}
\usepackage{algorithmic}
\usepackage{amsthm}
\usepackage{multirow}
\usepackage{pgffor}
\usepackage{graphicx}
\usepackage{epstopdf}
\usepackage{tikz}
\usepackage{epstopdf}
\usepackage{mathtools}
\usepackage{enumitem}
\usepackage{subfigure}
\usepackage{tabulary}
\usepackage{color}
\usepackage{xcolor}
\usepackage{url}
\usepackage{flushend}
\usepackage{stfloats}
\usepackage{tcolorbox}
\usepackage[utf8]{inputenc}
\usepackage[english]{babel}
\usepackage{tabularx}
\usepackage{CJKutf8}
\usepackage{microtype}
\usepackage{textcomp}
\usepackage{booktabs}
\usepackage{colortbl}
\usepackage{soul}
\usepackage{arydshln}
\usepackage{natbib}
\usepackage{appendix}

\usepackage{xspace}
\usepackage{microtype}
\usepackage{balance}
\usepackage{flushend}
\usepackage{multirow}
\usepackage{mathrsfs}
\newcommand{\cmark}{\ding{51}}%
\newcommand{\xmark}{\ding{55}}%
\usepackage[utf8]{inputenc}

\newcommand{\gptneox}{\textsc{GPT-NeoX}\xspace}
\newcommand{\codegen}{\textsc{CodeGen}\xspace}
\newcommand{\codex}{\textsc{Codex}\xspace}
\newcommand{\codet}{\textsc{CodeT5}\xspace}
\newcommand{\rine}{\textsc{rine}\xspace}
\newcommand{\knntop}{\textsc{$k$NN-ICL}\xspace}
\newcommand{\knnlm}{\textsc{$k$NN-LM}\xspace}
\newcommand{\icl}{\textsc{ICL}\xspace}
\newcommand{\finetune}{\textsc{Fine-Tune}\xspace}

\title{ \knntop: Compositional Task-Oriented Parsing Generalization with Nearest Neighbor In-Context Learning}

\author{
Wenting Zhao$^1$~~~~ Ye Liu$^2$~~~~ Yao Wan$^3$~~~~ Yibo Wang$^1$ ~~~~ Qingyang Wu$^4$  \\
  \textbf{Zhongfen Deng}$^1$~~~~ \textbf{Jiangshu Du}$^1$~~~~ \textbf{Shuaiqi Liu}$^5$~~~~ \textbf{Yunlong Xu}$^6$~~~~ \textbf{Philip S. Yu}$^1$\\ 
  $^1$University of Illinois Chicago ~~~~ $^2$Salesforce Research\\
  $^3$Huazhong University of Science and Technology ~~~~ $^4$Columbia University \\
  $5$The Hong Kong Polytechnic University ~~~~ $^6$State University of New York at Binghamton \\
  \texttt{wzhao41@uic.edu}
}

\begin{document}
\maketitle
\begin{abstract}
\textit{Task-Oriented Parsing (TOP)}  enables conversational assistants to interpret user commands expressed in natural language, transforming them into structured outputs that combine elements of both natural language and intent/slot tags. 
Recently, \textit{Large Language Models (LLMs)} have achieved impressive performance in synthesizing computer programs based on a natural-language prompt, mitigating the gap between natural language and structured programs. 
Our paper focuses on harnessing the capabilities of LLMs for semantic parsing tasks, addressing the following three key research questions: 1) \textit{How can LLMs be effectively utilized for semantic parsing tasks?} 2) \textit{What defines an effective prompt?} and 3) \textit{How can LLM overcome the length constraint and streamline prompt design by including all examples as prompts?}  
We introduce $k$ Nearest Neighbor In-Context Learning (\knntop), which simplifies prompt engineering by allowing it to be built on top of any design strategy while providing access to all demo examples.
Extensive experiments show that: 
1) Simple ICL without $k$NN search can achieve a comparable performance with strong supervised models on the TOP tasks, and 
2) \knntop significantly improves the comprehension of complex requests by seamlessly integrating ICL with a nearest-neighbor approach. Notably, this enhancement is achieved without the need for additional data or specialized prompts.
\end{abstract}

\section{Introduction}
\begin{figure}[t]
    \centering
    \includegraphics[width=0.98\textwidth/2]{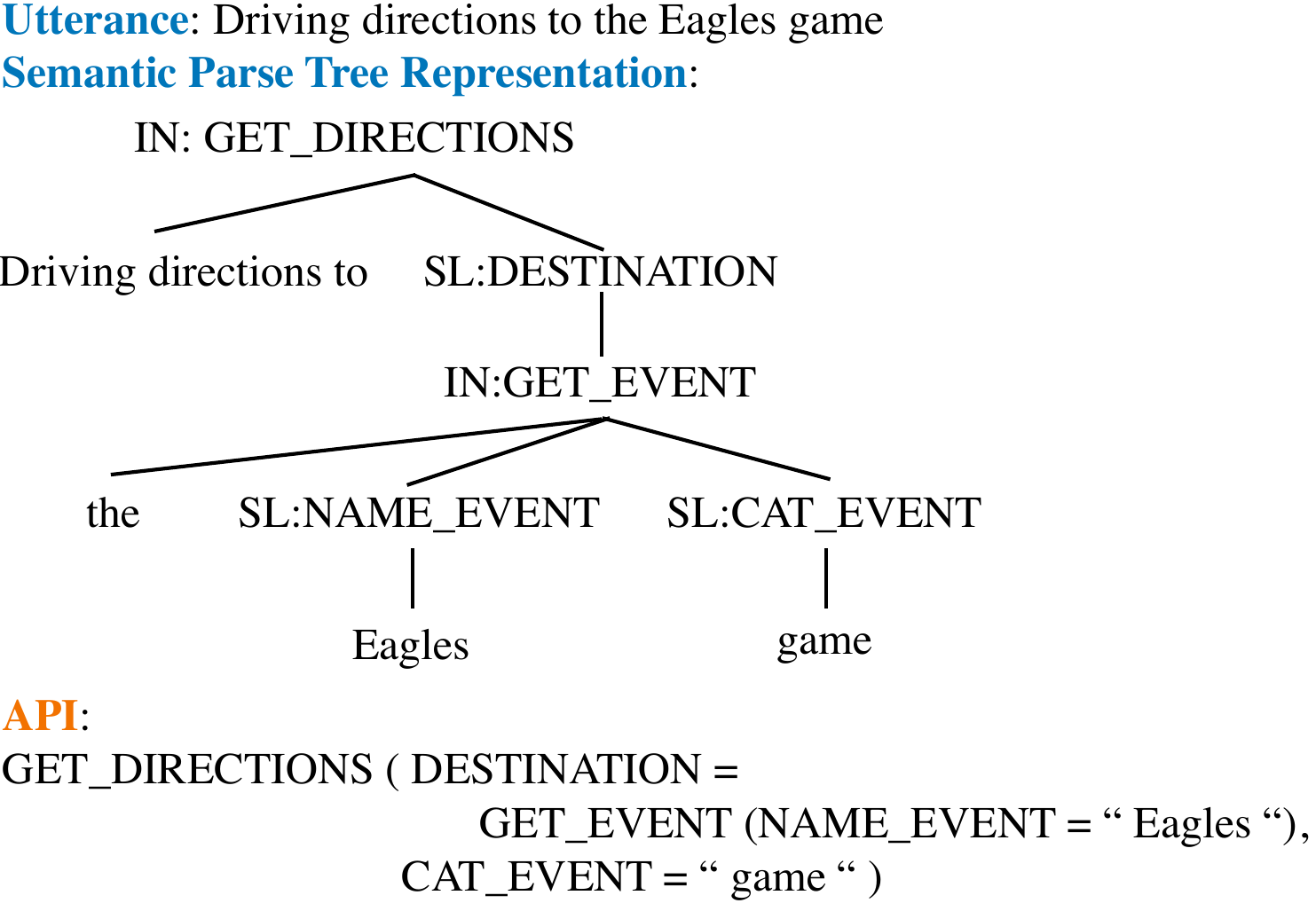}
    \caption{Reduction of an utterance from Topv2 with semantic parse tree to Python style API.}
    \label{fig:code_reduction}
    \vspace{-1em}
\end{figure}

\textit{Task-Oriented Parsing (TOP)}, which aims to translate the natural-language commands from users into specific actions, such as booking a restaurant, is an essential component in conversational systems~\citep{chen2020low,budzianowski-etal-2018-multiwoz,wu-etal-2023-diacttod}.
Over the past few years, a wide range of methods have been developed, ranging from rule-based methods~\cite{zelle1996learning,dong2018coarse} to neural program synthesis methods~\cite{shin2019program,mansimov2022semantic,drozdov2022compositional}.

The recent advancements in TOP involve formulating the task as a sequence-to-sequence problem, relying on a wealth of labeled data. This process typically entails feeding user utterances into Transformer models~\cite{liu2019roberta,devlin2018bert,raffel2020exploring} and designing specialized Transformer architectures to generate structured outputs that combine natural language with intent and slot tags like "IN:" and "SL:".
Some alternative approaches attempt to reframe TOP as conventional tasks, such as canonical paraphrasing~\cite{shin-etal-2021-constrained} and abstractive question answering~\cite{zhao-etal-2022-compositional}. These approaches aim to reduce the model's burden by eliminating the need to rationalize non-linguistic labels that were not part of pre-training.
However, in real-world scenarios, obtaining high-quality training examples with expert annotations can be challenging. Additionally, API documentation, which contains valuable information, often remains underutilized within the fine-tuning paradigm.
 


On the contrary, \textit{Large Language Models (LLMs)} excel in challenging few-shot scenarios, where they only need a few examples to generate the desired output and can also understand extensive API documentation. In this paper, we delve into an examination and analysis of the performance of ICL in the context of TOP. Besides, following a comprehensive review of prompt design strategies, we introduce \knntop, which offers adaptable integration with any prompt design strategy, further augmenting model performance.

We conducted experiments using three representative models: GPT-Neox-20B~\cite{black2022gpt}, CodeGen-16B-Multi~\cite{shin2019program}, and Codex~\cite{chen2021evaluating} code-davinci-002 to assess the performance of LLMs in TOP. Our research revolves around three key questions:

\textbf{\underline{Question 1}: How can we effectively leverage LLMs for TOP tasks?} We transform TOP into a code generation task, mapping semantic parse trees to Python code (referred to as `API' hereafter) to align with LLMs' output format, as shown in Figure~\ref{fig:code_reduction}.

\textbf{\underline{Question 2}: What constitutes proper prompt design for TOP using LLMs? } We analyze ICL performance under various prompt design strategies, including factors: API documentation and exemplar selection methods. Our findings include:
API documentation benefits the more powerful \codex model but can be distracting for the less-capable \codegen and \gptneox models.
Unsupervised demo selection proves most effective with ICL, thanks to its flexibility in capturing semantically similar examples holistically.

\textbf{\underline{Question 3}: How can we overcome the LLM's length constraint and simplify prompt design by including all examples as prompts?}
Instead of solely focusing on LLM performance through complex prompt design strategies, we draw inspiration from the retrieval language model approach~\cite{khandelwal2019generalization,khandelwal2021nearest} and introduce \knntop. \knntop enables LLMs to access all available demo exemplars during inference, harnessing synergy between the copy mechanism and the labeling task.
We tackle two key challenges.
First, how to infuse TOP knowledge into LLMs, considering their limited domain expertise due to the scarcity of labeled data? We address this by including a few demos as \knntop prompts, guiding LLMs to follow the desired output pattern.
Second, how to retrieve the nearest neighbor when there's a representation discrepancy between the \knntop datastore and the LLM prompt? To address this challenge, instead of using the hidden state $h_t$ directly from the LLM as the current step representation, we utilize a combination of the target utterance and previously generated words from the LLM at time step $t$. This ensures consistency in representation between the \knntop datastore and LLMs, offering flexibility in adapting to various prompt design strategies.

In summary, our contributions can be distilled into three key aspects:
1). We examine prompt design strategies for LLMs in the context of TOP by framing TOP as a code generation task. We conclude that similarity-based demo selection is effective for both black-box and open-source models, with stronger LLMs benefiting more from documentation;
2). We introduce \knntop to address the length constraints of LLMs, offering flexibility to integrate with any prompt design strategy;
and 3). Extensive experimental results showcase that \knntop can enhance the generation of better-nested API structures by leveraging guidance from nearest neighbors.
\section{Methodology}

\begin{figure*}[t]
    \centering
    \includegraphics[width=0.98\textwidth]
    {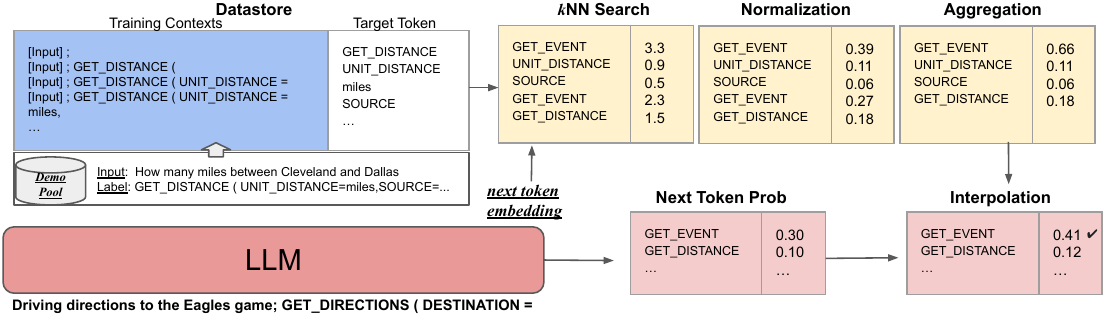}

    \caption{An illustration of \knntop at decoding time step $t$. The input to LLM is the user utterance ``\textit{Driving directions to the Eagles game}'' combined with the previously generated output ``\textit{GET\_DIRECTIONS ( DESTINATION =}''. The target token ``\textit{GET\_EVENT}'' is generated as output.
    }
    \label{fig:knnicl}
    \vspace{-1em}
\end{figure*}

In this section, we outline the methodology employed in this study. Our approach consists of two key components:
(1) Prompt Design for Semantic Parsing (Section~\ref{section_prompt_design}): We begin by crafting an effective prompt for TOP. This involves varying two crucial elements within the prompt: API documentation and exemplar selection strategies.
(2) Integration with \knntop (Section~\ref{section_knnicl}): we then integrate all the exemplars to LLM in \knntop. This integration allows us to harness the collective knowledge from all exemplars within the demo pool, enhancing the generation of the semantic parse API.

\subsection{Preliminaries}

\textit{In-Context Learning (ICL)}, which involves no parameter updates or fine-tuning for downstream tasks, has demonstrated competitiveness across multiple \textit{Natural Language Understanding (NLU)} and \textit{Natural Language Generation (NLG)} tasks. Typically, prompts are designed as task instructions, often comprising a few exemplars paired with input-output examples retrieved from a demo pool. However, due to the limited input length of LLMs, the number of exemplars is constrained.

More recently, the concept of retrieval language models~\cite {khandelwal2019generalization,pmlr-v162-borgeaud22a} has emerged. In this paradigm, language models have access to extensive corpora and employ explicit memory mechanisms for text generation. This approach has proven to enhance the capabilities of language models significantly.

\subsection{Reduction to Code Generation}\label{section_code_generation}
We utilize pre-trained code style knowledge, as found in models like \codex, \codegen, and \gptneox, trained on extensive code style corpora. We transform the API representations into Python code style, mapping the root node's intent name from the TOP semantic parse tree to the outermost Python function name. The tree branches serve as variable-value pairs within the Python function structure, enabling effective use of pre-trained code style knowledge.

\subsection{Prompt Design for Semantic Parsing}\label{section_prompt_design}
The fundamental prompt design for TOP consists of three core elements: ``\texttt{[API documentation]+[exemplars]+[target utterance]}.'' The target utterance represents the expected prediction by the LLMs, based on the provided prompt. We will delve into the first two components in the subsequent discussion.

\subsubsection{API Documentation}
In each domain, the API name, which includes intent and slot names from the TOP dataset, is represented symbolically, like \textit{GET\_LOCATION}. The API document or description offers natural language explanations of the domain service, aiding pre-trained LLMs in comprehension.

\subsubsection{Demo Selection}
We explore three exemplar selection strategies.

\noindent \textbf{Random Selection}
Exemplars are randomly drawn from the demo pool to create the prompt.

\noindent \textbf{Unsupervised Selection with SentenceBERT}
Utilizing SentenceBERT, we generate sentence-level embeddings for both the target utterance and all exemplar utterances. We rank the cosine similarity scores for all demos in the pool and select the top-$k$ most similar exemplars. In experiments, we set the number of exemplars ($m$) to 10 for all three models.

\noindent \textbf{Supervised Selection with Paraphrasing}
In contrast to SentenceBERT, we train a more precise classifier to rank utterances based on the similarity at the outermost intent level. We frame the similarity ranking as a paraphrasing task. To construct pairwise data for the paraphrase model, we label \texttt{[a, b]} as \texttt{True} if \texttt{a} and \texttt{b} share the same intent name from a parent node of the semantic parse tree, where \texttt{b} is sampled from the demo pool. For each utterance \texttt{a}, the ratio of constructed \texttt{True} pairs to \texttt{False} pairs is 1 to 5. During inference, we use the probability of predicting the label \texttt{True} as the ranking score.

\subsection{$k$-Nearest Neighbour In-Context Learning}\label{section_knnicl}


LLMs face a limitation in fully utilizing the exemplar information from the demo pool, making the demo selection strategy a critical factor influencing generation quality for ICL. Here, we introduce a comprehensive model, \knntop, designed to enable ICL to consider all demo exemplars from the demo pool, irrespective of the length constraint imposed by LLMs.

An Overview of \knntop is shown in Figure~\ref{fig:knnicl}: \knntop leverages both LLMs' predictions and $k$ nearest neighbor search to generate the final prediction. This involves probability distribution over the LLMs' vocabulary during the decoding process and the retrieval of the top-k nearest neighbors from the datastore.

A challenge for \knntop arises from the need to copy a span from the target utterance as a slot value, a requirement of the TOP task. Unfortunately, traditional $k$NN-LM~\citep{khandelwal2019generalization} or $k$NN-MT~\citep{khandelwal2020nearest} methods do not naturally support this. 
Furthermore, the constrained size of the demonstration pool may not encompass all potential slot values during the inference process. To tackle these challenges, we undertake the following steps:

\paragraph{LLM prompt}
\knntop bridges the information formality gap between LLM and the datastore. ICL has demonstrated promising copying ability from target utterances. Hence, we prompt the LLMs with a combination of exemplars and the target utterance to compensate for the limited datastore records, ensuring that the slot value is grounded in the target utterance.

\paragraph{Datastore Creation}
We create a datastore offline, comprising multiple key-value pairs. The key represents the contextualized representation of the input sentence encoded by LLM, and the value corresponds to the subsequent token of the input sentence. 
Given a set of training contexts denoted as $C$ and a target represented as $T$, the formulation of the datastore $D$ is as follows:
\begin{equation}
\begin{split}
D & \overset{\text{def}}{=} (\text{K}, \text{V}) = \{(f(c_i), t_i) | \\
 & \forall c_i \in C, \forall t_i \in T,  (c_i, t_i) \in (C, T) \} 
\end{split}
\end{equation}
For each example in the training set, we store multiple datastore records containing only tokens corresponding to the semantic parsing labels as targets.

\paragraph{Similarity Search}
Another challenge arises because the hidden state produced by LLM at time step $t$ resides in a different representation space than the datastore. To address this, at time step $t$ when generating token $y_t$, we calculate the current hidden state representation as a combination of the target utterance $s$ and the previously generated API tokens $y_{1 \dots i-1}$. This alignment ensures compatibility with the datastore representation space. Given the limited size of the datastore in TOP, the distribution of retrieved $k$ nearest neighbors tends to be skewed. To mitigate overfitting to the most similar retrieval, we introduce a temperature parameter $Temp$ to flatten the distribution.
\begin{multline}
p_{kNN} (y_t|c, y_{1:t-1}) \propto \\
\sum_{(k_j,v_j) \in N}1_{y_t=v_j}\exp\left(\frac{-Dis(k_j,f(c,y_{1:t-1}))}{Temp}\right)
\end{multline}

\paragraph{Interpolation}
The decoding process involves interpolating the $k$ nearest neighbor from the datastore with the language model distribution.
To enable the model to accurately predict slot values by copying spans from the target utterance, we employ the full vocabulary from LLM rather than restricting it to the intersection between LLM vocabulary and the $k$ nearest neighbor vocabulary.
\begin{multline}
p(y_t|x,y_{1:t-1})=\lambda p_{kNN} (y_t|c, y_{1:t-1}) \\
+ (1-\lambda)p_{lm}(y_t|x,y_{1:t-1})
\end{multline}

\noindent \textbf{\knntop vs. $k$NN-MT} \knntop is a generalization of $k$NN-MT for conditional generation tasks using In-Context Learning (ICL), seamlessly integrating with LLMs. It introduces some significant differences compared to $k$NN-MT. First, the LLM conditions not only on the source sequence but also on the provided prompt, which typically includes a few selected demonstrations. This enables the LLM to align more closely with the pattern of the target demonstration. Second, while the datastore in $k$NN-MT is constructed solely from the source and target mappings, the LLM produces representations that are conditioned on the entire prompt and target. This results in a discrepancy when selecting the k nearest neighbors. To address this, we rely on representations derived from both the source and the previously generated tokens in the target to ensure consistency during similarity search.
\section{Experiments}
In this section, we address the following key research questions through our experiments: (1) What constitutes an effective prompt design to enhance the performance of LLMs? (2) How does ICL compare to state-of-the-art supervised models on TOP tasks? (3) What impact does \knntop, which incorporates all available demo examples from the demo pool through interpolation, have on prediction quality?

\begin{table}[t]
\centering
\resizebox{0.98\textwidth/2}{!}{
\begin{tabular}{clcccc}
\toprule
\textbf{Exemplars} & \textbf{Doc}  & \textbf{\gptneox} & \textbf{\codegen} & \textbf{\codex} \\
\midrule
Random & \xmark & 2.02 & 3.54 & 19.20 \\
Unsupervised & \xmark & \textbf{6.23} & \textbf{10.27} & 36.03 \\
Supervised & \xmark & 4.04 & 9.60 & \textbf{37.54} \\
\midrule
Random &  \cmark & 1.68 & 3.37 & 26.43 \\
Unsupervised & \cmark & 4.04 & 7.91 & 39.23 \\
Supervised & \cmark & \textbf{4.04} & \textbf{9.43} & \textbf{41.25} \\
\bottomrule
\end{tabular}
}
\caption{\label{tab:ICL-prompt-sota}
Exact Match results for TOPv2 dataset on Navigation domain using LLMs with varying prompt components. 
}
\end{table}

 

\subsection{Dataset, LLMs, and Evaluation}
We conduct experiments on the TOPv2 dataset~\citep{chen2020low}, an extension of the TOP~\citep{gupta-etal-2018-semantic-parsing} dataset designed for voice assistants. TOPv2 encompasses 8 domains, including 6 new additions, providing a diverse range of task-oriented semantic parsing examples. Due to the resource-intensive nature of running LLMs, we select 4 representative domains for our testing: Navigation, Reminder, Alarm, and Weather. These domains were chosen based on their complexity, with Navigation and Reminder representing more intricate domains with complex nested semantic parses and a larger number of intent slot names, while Alarm and Weather offer relatively simpler examples with fewer complexities in terms of intent and slot names.

For our experiments, we employ GPT-Neox-20B~\cite{black2022gpt}, CodeGen-16B-Multi~\cite{shin2019program}, and Codex~\cite{chen2021evaluating} (\texttt{code-davinci-002}) to assess the performance of LLMs in ICL. GPT-Neox and CodeGen models are publicly available and can be hosted locally, whereas Codex is accessible solely through a commercial API provided by OpenAI. Codex cannot be downloaded and implemented directly with our proposed method \knntop. Hence, we employ CodeGen and GPT-Neox for experimenting with \knntop.

We evaluate model performance using exact match criteria, where a prediction is considered correct (scored as 1) if it matches the ground-truth precisely and incorrect (scored as 0) if there is any discrepancy between the prediction and the ground-truth. This evaluation metric disregards the order of the semantic parse tree branches, focusing solely on the correctness of the prediction.

\subsection{Baselines}
We consider three sets of baselines. \textbf{Supervised State-of-the-Art Models}: To establish a performance benchmark for LLMs on TOP, we include supervised state-of-the-art models. RINE~\citep{mansimov2022semantic} introduces a recursion insertion-based encoder tailored for TOP, breaking the task into smaller steps where each step predicts either an intent or slot name along with its starting and ending positions. CodeT5~\citep{wang2021codet5}, a unified pre-trained encoder-decoder model based on T5, showcases strong performance in code generation tasks. \textbf{ICL}: Unless specified otherwise, we will use ICL with randomly selected demos as the prompt. \textbf{$k$NN-LM}: In this baseline, we use $k$NN-LM without the inclusion of exemplar prompts for LLMs.

\begin{table*}[t]
\centering
\resizebox{0.75\textwidth}{!}{
\begin{tabular}{llccccc} 
\toprule

\textbf{LLM} & \textbf{Method}  & \textbf{Navigation} & \textbf{Reminder} & \textbf{Alarm} & \textbf{Weather} & \textbf{Avg} \\
\midrule
\codet &  \multirow{2}{*}{\finetune}   & 10.02  & 6.61  & 15.72 & 6.60  & 9.74\\
\rine &  &  42.28 & 36.87 & 32.09 &  32.53 & 35.94\\ 

\midrule
\multirow{3}{*}{\gptneox} 
& ICL & 1.81  & 5.28  & 9.54  & 16.56  &  8.30\\
& \knnlm & 1.22  & 3.48 & 6.54 & 2.28 &  3.38\\
& \knntop & \textbf{5.69}  &  \textbf{8.48} & \textbf{19.40} & \textbf{24.52}  &  \textbf{14.52} \\
\midrule
\multirow{3}{*}{\codegen} 
& ICL & 3.94 & 5.99  & 10.88 &   13.84 & 8.66 \\
& \knnlm & 0.51  & 0.14 &5.07  &  1.01 & 1.68 \\
& \knntop & \textbf{8.37}  & \textbf{10.49} & \textbf{19.10}   &  \textbf{25.19}   & \textbf{15.79} \\
\midrule
\multirow{2}{*}{\codex}
& ICL & 18.78 & 30.46  & 45.08 &  45.70 & 35.01 \\
& \knntop$^{*}$ & \textbf{35.74} & \textbf{41.36} & \textbf{57.56} & \textbf{53.35} & \textbf{47.00} \\

\bottomrule
\end{tabular}
}
\caption{\label{tab:spis10-knnicl}
Results of Exact Match on SIPS10 demo pool. We selected four domains using three LLMs: \gptneox, \codegen, and \codex on TOP. ICL refers to vanilla LM using randomly selected examples as demo; \knnlm uses the external dataset as retrieval pool while no demo is used as prompt; and \knntop is the proposed method using retrieved demo and external datastore. $^{*}$ indicates an estimated result from the black-box \codex model using retrieved semantic similar demo examples.}
\vspace{-1em}
\end{table*}

\subsection{Implementation Details}
For ICL, we utilize 10 exemplars to construct the prompt for LLMs. In the case of \knntop, we explore a range of parameter values, including temperature (chosen from {50, 100, 200, 300, 400, 500}), interpolation weight (selected from {0.1, 0.3, 0.5, 0.7}), and the number of neighbors (picked from {20, 100, 1000}). We conduct an exhaustive search for the best parameter combinations within each domain. We use FAISS~\citep{johnson2019billion}, an open-source library for fast nearest-neighbor retrieval in high dimensional spaces.

Our experiments are conducted on machines equipped with 8 Tesla V100 GPUs, each with 16GB of memory. In the context of \knntop, the inference batch size is set to 3 for \codegen and 1 for \gptneox, respectively.

To assess \knntop's performance, we establish the SPIS10 data split for each domain, where each domain comprises a maximum of 10 examples for intent or slot labels, simulating a few-shot setting. Additionally, we evaluate \knntop on a larger data setting with a demo pool containing 2000 randomly sampled examples.

\subsection{Prompt Design Results (RQ1)}
Table~\ref{tab:ICL-prompt-sota} presents the exact match scores achieved through the ablation of different prompt components, including the presence of API documentation and the use of three exemplar selection strategies. Due to resource constraints, we randomly selected 20\% of the validation data from the Navigation domain, resulting in 596 examples. Additionally, we include the results obtained using RINE and CodeT5 to provide a benchmark for ICL by comparing them with fine-tuning state-of-the-art results.

To summarize, first, the best performance for GPT-NeoX and CodeGen is achieved with prompts that do not include API documentation and utilize similarity-based selection strategies. This suggests that these two models struggle to leverage the documentation due to the input sequence length limitations, making it challenging to extract useful information. However, Codex benefits from API documentation when combined with the SentenceBERT similarity strategy. This difference may be attributed to Codex's stronger capacity to handle longer prompts and effectively utilize API documentation. 

When comparing the SentenceBERT and paraphrase similarity-based selection strategies, SentenceBERT yields a better prompt when documentation is not included. The paraphrased model provides higher-quality exemplars in terms of outermost intent name similarity, 
but it falls short of capturing the nested semantic parse tree structure, which is crucial for predicting the correct API.

\subsection{ICL vs. Supervised Methods (RQ2)}
Table~\ref{tab:spis10-knnicl} reports the results for supervised models: \rine and \codet on the few-shot setting. We create SPIS10 data split for each domain in which each domain contains at most 10 examples for intent or slot label. 

We compare ICL with fine-tuned SOTA models. Since \codex is a black-box model, we use the top-k most similar examples as prompt to estimate its \knntop performance, as shown in Table~\ref{tab:spis10-knnicl}.
We observed that the \codex model consistently outperforms \rine by an average margin of 11.06 across four domains, with notable improvements of 4.5\%, 25.5\%, and 20.8\% in the Reminder, Alarm, and Weather domains, respectively.
However, \rine shows better performance on the most challenging domain Navigation. To summarize, the \codex model demonstrates superior performance over SOTA models especially on flattener domains, while slightly under-performs on the domain with more challenging examples. 
Although \gptneox and \codegen models lag behind \codet and \rine model, \codex results show the potential of ICL from the high capacity model.


\subsection{\knntop Results  (RQ3)} 
\noindent \textbf{Few-shot setting} 
In Table~\ref{tab:spis10-knnicl}, we showcase the results for our novel approach, \knntop, alongside the baseline method, \knnlm.
From this table, we can observe that \knntop outperforms \knnlm across all domains, demonstrating its effectiveness in leveraging prompts for TOP. Notably, \knntop achieves improvements of 11.1\% and 14.1\% when compared to \knnlm, using \gptneox and \codegen, respectively.

We also compare \knntop with its \icl counterpart. The results show that for \gptneox and \codegen, \knntop provides a boost in performance across an average of four domains, with gains of 6.22\% and 7.13\% in exact match scores. It's worth noting that our datastore size is limited, containing approximately 100 examples, which matches the size of the demo pool. Therefore, \knntop effectively leverages all available example information to achieve these improved results.

\begin{table*}[t]
\centering
\resizebox{0.75\textwidth}{!}{
\begin{tabular}{llccccc} 
\toprule
\textbf{LLM} & \textbf{Method}  & \textbf{Navigation} & \textbf{Reminder} & \textbf{Alarm} & \textbf{Weather} & \textbf{average} \\
\midrule
\multirow{3}{*}{\codegen}
& ICL-Retrieve & 37.00  & 23.40  & 48.10 &  65.80  & 43.58 \\
& \knnlm & 21.30 & 3.10  &  26.99 &   46.60 & 24.50 \\
& \knntop & \textbf{38.20} & \textbf{24.20} & \textbf{50.40} &  \textbf{66.60}   &  \textbf{44.85}\\
\midrule
\multirow{3}{*}{\gptneox} 
& ICL-Retrieve & \textbf{33.40} & \textbf{24.10}  & 46.70 & 65.70  &  42.48\\
& \knnlm & 14.30 &  6.70 & 33.50 & 40.00  & 23.63 \\
& \knntop & 33.30  &  24.00 &  \textbf{46.90} &   \textbf{66.30} &  \textbf{42.63}\\
\bottomrule
\end{tabular}
}
\caption{\label{tab:2000-knnicl}
Results for Exact Match on a randomly sampled pool of 2000 demonstrations, illustrating the scalability of \knntop.
}
\end{table*}

\noindent \textbf{Scale Up} We conducted experiments to test the performance of \knntop on a larger datastore, which includes 2000 examples. The goal was to determine if \knntop could benefit more from a larger datastore compared to \knnlm. 
When comparing \knntop with \knnlm, we found that the performance gap between the two models becomes larger in the presence of the larger datastore. Notably, \knntop outperforms \knnlm by 20.3\% and 19.0\% using \codegen and \gptneox as backbone models respectively.
Furthermore, we compared \knntop to ICL-Retrieve, a method that retrieves the most semantically similar example from the demo. In this comparison, \knntop outperforms \codegen by 1.2\%, 0.8\%, 2.3\%, and 0.8\% on the Navigation, Reminder, Alarm, and Weather domains, respectively. This improvement highlights that even when a demo selection strategy is fixed, \knntop can still bring about performance enhancements.

For \gptneox, the \knntop shows similar performance on the Navigation and Reminder domains, 
with a slight improvement observed in the Alarm and Weather domain.

\begin{table}
\centering
\setlength{\tabcolsep}{6pt} 
\resizebox{0.8\textwidth/2}{!}{
\begin{tabular}{c|cc|cc}
\toprule
\multirow{3}{*}{Depth} & \multicolumn{2}{c}{\textbf{\gptneox}} & \multicolumn{2}{c}{\textbf{\codegen}} \\

\cmidrule(lr){2-3} \cmidrule(lr){4-5} 
 &  ICL & \knntop & ICL & \knntop \\
\midrule
1  & 2.53 & \textbf{6.42} & 5.90 & \textbf{10.16} \\
2  & 1.14 & \textbf{5.72} & 1.72 & \textbf{6.45} \\
3  & 0.00 & \textbf{1.93} & 0.16 & \textbf{4.98} \\
\bottomrule
\end{tabular}
}
\caption{Analysis of accuracy with respect to nesting depth (1/2/3) for the TOPv2 dataset in the Navigation domain using different models.}
\label{tab:depth}
\end{table}

\begin{figure}[t]
    \centering
    \includegraphics[width=0.98\textwidth/2]{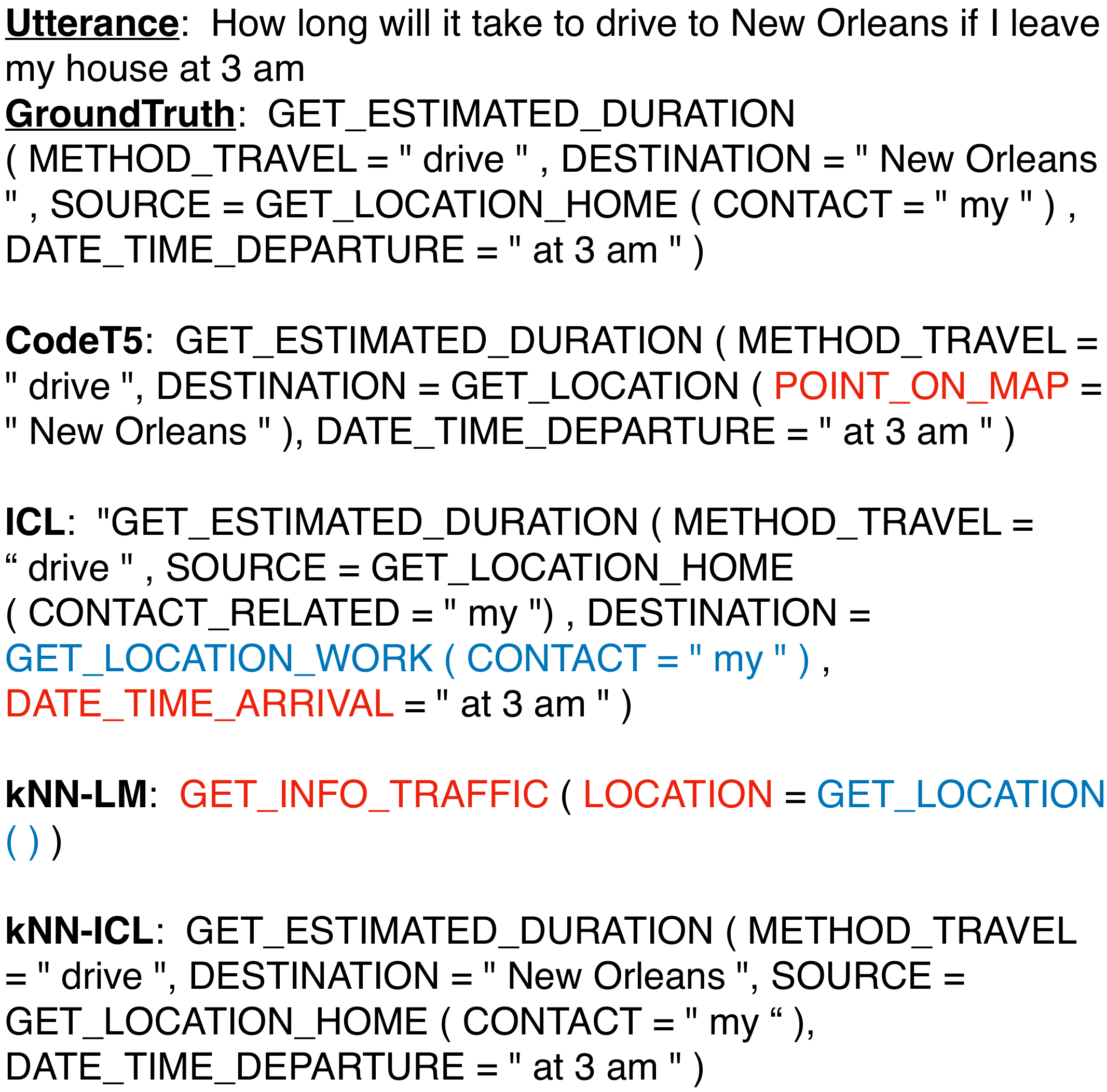}
    \caption{A case study from Topv2. The wrong intent and slot name from prediction are marked in red, and the hallucinated nested API call is marked in blue. The ICL, \knntop, \knnlm uses \gptneox model.}
    \label{fig:case_study}
    \vspace{-1em}
\end{figure}
\subsection{Additional Analysis: Depth vs. Accuracy}
Table~\ref{tab:depth} illustrates the model's performance concerning the semantic depth in the Topv2 dataset. Our analysis focuses on the Navigation domain, which contains the most complex average semantic parse tree depth. 
Notably, ICL on \gptneox and \codegen exhibits a consistent performance drop with increasing depth. 
On the other hand, \knntop significantly enhances the performance of examples compared to its ICL counterpart. 


\subsection{Case Study}
In order to assess the performance of state-of-the-art models, specifically ICL, \knnlm, and \knntop, we present a case study in Figure~\ref{fig:case_study}. This case study provides examples of utterances, corresponding ground-truth APIs, and model predictions. The showcased example includes a nested semantic parse tree structure with a depth of 3.

First, \codet exhibits overall good quality, except for hallucinating the slot name \texttt{POINT\_ON\_MAP} on the deepest tree branch. However, \icl had a hard time generating API calls for deeper-level APIs, such as hallucinating \texttt{GET\_LOCATION\_WORK ( CONTACT = " my " )} for slot \texttt{DESTINATION}. \knnlm suffers from the wrong outermost intent name prediction as well as the slot name prediction because no prompt is provided, causing severe deviation from the expected output. In contrast, \knntop performs best on this complex example with the aid of both prompting, producing an accurate utterance span as a slot value, and the $k$ nearest neighbor guidance, choosing the proper nested API structure.

\section{Related Work}


\paragraph{Semantic Parsing}
Semantic parsing plays a vital role in developing commercial personal assistants for understanding spoken language. Supervised models, including sequence-to-sequence models~\citep{mesnil2013investigation, liu2016attention} and pretrained language models~\citep{devlin2018bert, chen2019bert}, have demonstrated competitive performance. In structured semantic parsing, recent approaches break down semantic tree construction into multiple steps. For instance, RINE~\citep{mansimov2022semantic} recursively inserts predicted labels at predicted positions to construct the semantic parse tree. Additionally, some methods, like~\citep{zhao-etal-2022-compositional}, transform semantic parsing into multi-turn abstractive question answering. In few-shot scenarios, techniques such as meta-learning~\citep{chen2020low} and label semantics~\citep{paolini2021structured, desai2021low} enhance neural semantic parsers. Furthermore,~\citep{shu2022dialog2api} simplifies semantic parsing to program synthesis and integrates the predicted program into an executable environment.

\paragraph{In-Context Learning}
With the emergence of Large Language Models~\citep{nijkamp2022conversational, black2022gpt, brown2020language, scao2022bloom, chen2021evaluating, chowdhery2022palm}, model parameters have scaled from 16 billion to over 500 billion. The substantial resource requirements for hosting LLMs have made it challenging to fine-tune them for downstream tasks. As a result, the paradigm for utilizing language models has shifted from traditional pre-training and fine-tuning to more efficient approaches, such as parameter-efficient fine-tuning\citep{lester2021power, li2021prefix, he2021towards} and in-context learning~\citep{dong2022survey, liu-etal-2022-makes}.
ICL has demonstrated its reasoning capabilities in both plain text~\citep{wei2022chain, zhang2022automatic} and structured text formats~\citep{chen2022large, chen2022program}. Researchers have also analyzed the factors contributing to the effectiveness of ICL~\citep{dai2022can, min2022rethinking, von2022transformers} in general tasks.
However, existing literature primarily focuses on the overall performance of ICL in generic tasks. Our work, on the other hand, falls within a specific domain, where we investigate the performance of ICL in the context of task-oriented semantic parsing problems.

\paragraph{Retrieval-Augmented Large Language Model}
Recent advancements in Retrieval-Augmented Language Models (RALMs) have garnered significant attention within the NLP community. LLMs face challenges in terms of scaling~\citep{khandelwal2019generalization} and acquiring long-tail knowledge~\citep{mallen-etal-2023-trust, izacard2022few}. RALMs~\citep{guu2020retrieval, liu2023exploring} present a promising solution by seamlessly integrating non-parametric data stores with their parametric counterparts. To enhance the adaptability of RALMs for downstream tasks, researchers have devised strategies such as zero or few-shot prompting~\citep{shi-etal-2022-nearest, xu2023k} and fine-tuning~\citep{shi2023replug}. To the best of our knowledge, our work stands as the pioneering effort to combine the $k$ nearest neighbor retrieval language model with in-context learning on generative tasks.

\paragraph{Code Generation} Automated code generation dates back as far as a few decades ago~\citep{backus1957fortran, manna1971toward}. Recently, the code generation has been dominated by the LLMs, including the closed-source models AlphaCode~\citep{li2022competition} and ChatGPT~\citep{ChatGPT}, as well the open-source models such as CodeT5~\citep{wang2021codet5}, CodeT5+~\citep{wang2023codet5+}, CodeGen~\citep{nijkamp2022codegen}, InCoder~\citep{fried2022incoder}, StarCoder~\citep{li2023starcoder}, and Code Llama~\citep{codellama}.
Inspired by this line of research, our work extends the scope of TOP to encompass code generation.
\section{Conclusion}

In this paper, we delved into the intricacies of prompt design and introduced \knntop to include all demo examples on the Task-Oriented semantic parsing task, addressing input length challenges during decoding and lessening prompt engineering efforts. Our findings highlight the substantial impact of prompt design on TOP, with models like \codex deriving more benefits from structured documentation than less capable models. The similarity-based exemplar retrieval strategy enhanced model performance significantly. Furthermore, \knntop's integration with a large language model and $k$ Nearest Neighbor search ensures comprehensive use of training data, offering more consistent results across tasks. This study underscores the critical role of prompt design, pointing to promising avenues for future research.

\section{Limitations}
\noindent \textbf{Generalization of Prompt Design}  This study highlights the varying impact of prompt design on different models based on their capacities. Instead of advocating a universal prompt design strategy, the optimal choice should be model-agnostic. A potential avenue for future research is enabling models to autonomously select their preferred prompt designs~\cite{zamfirescu2023johnny,zhou2022large}. Depending on the LLM's chosen strategy, \knntop can serve as a plugin to further enhance model performance by harnessing all available examples.

\noindent \textbf{Generalization to High-Capacity Models} This paper offers an initial exploration of \knntop with LLMs. While \codex is not publicly accessible and the resource limitation, we have used \gptneox and \codegen as representative models for \knntop experiments. Nevertheless, given the substantial performance disparity between \gptneox, \codegen, and \codex, future research should delve into assessing the influence of \knntop on more robust models.


\bibliographystyle{acl_natbib}
\bibliography{custom}




\end{document}